\documentclass[conference]{IEEEtran}
\IEEEoverridecommandlockouts
\IEEEpubid{\makebox[\columnwidth]{978-1-6654-3886-5/21/\$31.00 \textcopyright 2021 IEEE \hfill} \hspace{\columnsep}\makebox[\columnwidth]{ }}
\usepackage{cite}
\usepackage{amsmath,amssymb,amsfonts}
\usepackage{xurl}
\usepackage{algorithmic}
\usepackage{enumerate}
\usepackage{graphicx}
\usepackage{textcomp}
\usepackage{xcolor}
\usepackage{url}
\usepackage{units}
\usepackage{csquotes}
\def\BibTeX{{\rm B\kern-.05em{\sc i\kern-.025em b}\kern-.08em
    T\kern-.1667em\lower.7ex\hbox{E}\kern-.125emX}}

\newcommand{\set}[1]{\ensuremath\mathcal{#1}}
\newcommand{\vect}[1]{\ensuremath\mathbf{#1}}
\newcommand{\card}{\ensuremath\vect{c}}
\newcommand{\mycards}{\ensuremath\set{C}}
\newcommand{\pack}{\ensuremath\set{P}}
\newcommand{\pref}{\succ}

\begin{document}
\bibliography{bibliography.bib}

\title{Predicting Human Card Selection in \emph{Magic: The Gathering} with Contextual Preference Ranking}

\author{\IEEEauthorblockN{Timo Bertram}
\IEEEauthorblockA{\textit{Dept.\ of Computer Science}\\
\textit{Johannes-Kepler Universit\"at} \\
\textit{Linz, Austria} \\
\url{tbertram@faw.jku.at}}
\and
\IEEEauthorblockN{Johannes F\"urnkranz}
\IEEEauthorblockA{\textit{Dept.\ of Computer Science}\\
\textit{Johannes-Kepler Universit\"at} \\
\textit{Linz, Austria} \\
\url{juffi@faw.jku.at}}
\and
\IEEEauthorblockN{Martin M\"uller}
\IEEEauthorblockA{\textit{Dept.\ of Computing Science} \\
\textit{University of Alberta}\\
\textit{Edmonton, Canada} \\
\url{mmueller@ualberta.ca}}
}

\maketitle

\begin{abstract}
Drafting, i.e., the iterative, adversarial selection of a subset of items from a larger candidate set, is a key element of many games and related problems. It encompasses team formation in sports or e-sports, as well as deck selection in formats of many modern card games. The key difficulty of drafting is that it is typically not sufficient to simply evaluate each item in a vacuum and to select the best items. The evaluation of an item depends on the context of the set of items that were already selected earlier, as the value of a set is not just the sum of the values of its members - it must include a notion of how well items go together. 

In this paper, we study drafting in the context of the card game Magic: The Gathering. We propose the use of the Contextual Preference Ranking framework, which learns to compare two possible extensions of a given deck of cards. We demonstrate that the resulting neural network is better able to better inform decisions in this game than previous attempts.
\end{abstract}


\section{Introduction} 

Collectible card games have been around for decades and are among the most played tabletop games in existence. However, they are also among the most complex games \cite{b10}. Of course, a good player needs to be able to play the game itself, which requires an understanding and knowledge of thousands of cards. Furthermore, \textit{deck-building}, choosing a suitable set of cards to play with, is a gigantic challenge in itself. For the game of \emph{Magic: The Gathering}, a lower boundary of the number of possible card configurations can be computed as follows. For one of their most restricted game modes, \emph{Standard}, 1983 different cards are currently legal. Decks consist of at least 60 cards, of which usually about 37 are chosen from the aforementioned pool, which we will use as our lower bound. As each card can be put into a deck up to four times, this leads to ${1983 \times 4 \choose 37} > 10^{101}$ combinations of cards. Even with the assumption that a player will play a deck black and blue deck, which is a more reasonable assumption, this still results in ${847 \times 4 \choose 37} > 10^{87}$ possible decks.

As such numbers are vastly beyond the power of exhaustive computation, other methods must be developed to train agents to build decks. In this work, we study a specific game mode of MTG where the deckbuilding process is greatly simplified. To train and evaluate our method, we use a dataset of expert draft selections, which provides information about which selections human experts preferred over others. 

Our main technical contribution is to use Siamese networks in a way that has not been previously used. We train and employ them to decide between different choices by explicitly modeling the context of the decision. The general framework of this method of Contextual Preference Ranking is developed in Section \ref{sec:CPR}. Before that, we start with a brief description of the game (Section~\ref{sec:MTG}) and a review of related work and Siamese networks (Sections~\ref{sec:related} and~\ref{sec:Siamese}).
Our experiments and their results are presented in Sections~\ref{sec:experiments} and~\ref{sec:results}, followed by a discussion, our conclusions, and an outlook on open questions for future work.

\section{Magic: The Gathering}
\label{sec:MTG}

\emph{Magic: The Gathering (MTG)} is a collectible card game with several million players worldwide.
We abstain from explaining the complex rules \cite{b15}, as they are not necessary to understand the contribution of this work, but provide some background information in order to introduce the terminology used.

\subsection{Drafting}
\label{sec:Drafting}
MTG is played in a variety of different styles. For this work, we consider the format of \emph{drafting} in a game with eight players. In contrast to formats where decks are constructed separately from playing, drafting features a first game phase in which players form a pool of cards, from which they later build their decks. The pool of cards is chosen from semi-random selections of cards, so-called \emph{packs}. Each pack of \emph{MTG} cards consists of 15 different cards of four different rarities: eleven \emph{Common}, three \emph{Uncommon}, and one \emph{Rare} or \emph{Mythic} card.\footnote{Some packs contain an extra sixteenth card. However, such packs did not occur in our dataset.}  \emph{Rare} cards appear more frequently than \emph{Mythic} ones. Over the course of the whole draft, each player chooses a deck of 45 cards sequentially. Players get their cards by choosing from many packs as follows: Each of the eight players in a draft starts with a full pack of 15 cards, selects a single card from it, and passes the remaining 14 cards on to the next player. In the following rounds, players select from 14, 13, \ldots cards. This process continues until all 15 cards of the original packs are chosen. This process is repeated for an additional two packs, such that each player selects 45 cards in total. In each round, packs are passed around in the same direction. 

Drafting differs from free deckbuilding since players can not choose any existing card. Still, the computational complexity of this problem is enormous, as a single draft leads to $(15\cdot 14\cdot 13 \ldots)^3 > 2\times 10^{36}$ possible decks for each individual player. As there are 8 players but 15 cards per pack, players will see most packs of cards twice. This gives players additional information, such as which cards have been selected by the opponents in the last round. Such information is disregarded in our current work. We evaluate each pick only in the context of our current selection of cards, without taking information about the opponents' possible picks, future or past, into consideration. 

\subsection{Card Colors}
An important property of a card is its \emph{card colors}, which has a major impact on the composition of a good deck. 
Most cards are assigned a single color, but some cards have multiple colors, and a small subset of cards has no colors. While there are exceptions to this, most players' final decks will only use cards of two different colors. This means that previously selected cards, especially their color, strongly influence the selection of subsequent cards, in order to build a consistent deck. This also means that colorless cards can be valuable, as they can be used in any deck. On the other hand, using a multicolored card however requires having \textit{all} of its colors in the deck, which makes them much harder to incorporate.

\subsection{Deck Building}
Only about 23 of the 45 selected cards will be used in a player's final deck, so almost half of all chosen cards do not participate in the play phase.\footnote{The reason is that a legal deck only requires 40 cards, of which usually 17 are so-called \emph{basic lands}, which are not part of the drafting process. For more information about this, visit \url{https://magic.wizards.com/en/articles/archive/lo/basics-mana-2014-08-18}.} This opens strategic options such as making speculative picks and changing colors during the drafting phase. Players may also pick some strong cards without intending to play them, in order to deny the other players that card.

In actual games, the draft phase is followed by the play phase. In this work, instead of evaluating a drafting strategy directly by playing games with the resulting deck, we evaluate it by using a large database of human expert picks. This serves as the ground truth for which card is best in a specific situation. This has its limitations, as human choices are far from perfect and can be inconsistent, as well as having no information about the performance of the resulting decks. Still, this dataset is useful when trying to predict human decision-making in this context, and allows the study of draft picking independently of card play.

\section{Related work}
\label{sec:related}
Current work on selecting cards in the setting of collectible card games is limited due to available data. Most existing approaches either drastically reduce the complexity of the domain by choosing subsets of cards \cite{b11} or by using naive versions of games \cite{b13}. Evolutionary approaches are often used for deck building. However, computing the fitness of a deck is a difficult problem by itself. In practice, those approaches often use na\"ive game-specific heuristics to play games \cite{b11,b12}, which do not transfer to the context of real gameplay. A different way to circumvent the complexity of evaluating decks, which we follow here, is by training on expert decisions. \textsl{DraftSim} \cite{b1} is a large public domain simulator for human deck-building decisions, which provides an excellent basis for training. This dataset uses the eight-player drafting setting explained in Section \ref{sec:Drafting}. In their work, they also proposed several card selection methods \cite{b1}. Their best performing method is a deep neural network, which was trained to directly pick the best-fitting card in each round. The input of the network is a feature-based encoding of the current set of cards, while the output is a vector of real-valued scores, which rank all possible card choices. A card with the maximum score within the current selection $\pack$ is chosen.

In our work, we train a Siamese network \cite{b8,b9} for the task of drafting. These networks are often used in one-shot learning for image recognition \cite{b14,b9} and process multiple inputs sequentially in the same network (see Section \ref{sec:Siamese}). Siamese networks have also been used in preference learning two compare two examples of a similar item \cite{RankNet}. This idea can also be viewed as an extension of Tesauro's comparison training networks \cite{ComparisonTraining}. In his work, networks use pairwise comparisons without context, while we add the anchoring context sets.
In our work, we use Siamese networks differently: we compare two items with a context by embedding both inputs as well as the context in a representation space with the help of the network. To the best of our knowledge, this is a novel approach.

\section{Siamese networks for preference learning}
\label{sec:Siamese}

\begin{figure}
    \centering
    \includegraphics[width = 0.75\linewidth]{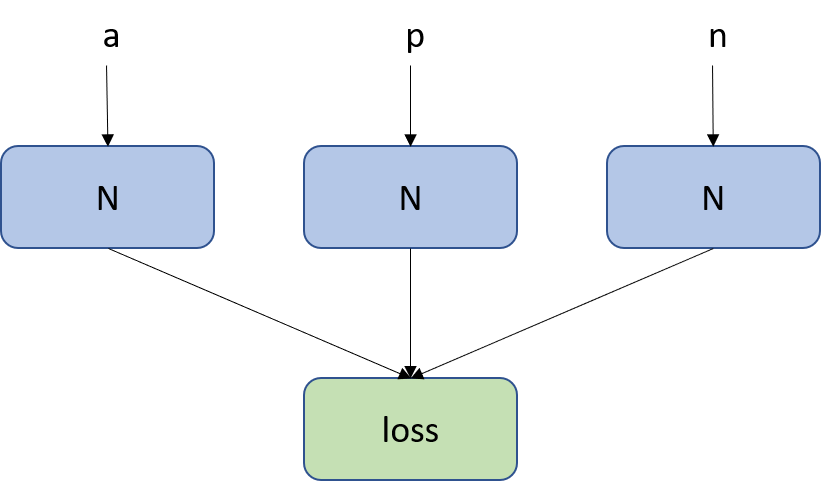}
    \caption{Training scheme for triplet loss using an anchor $\vect{a}$, a positive $\vect{p}$ and a negative example $\vect{n}$. The loss function indicates whether $\vect{a}$ is closer to $\vect{p}$ or to $\vect{n}$. $N$ is the network that maps an item into an embedding space.
    }
    \label{fig:scheme}
\end{figure}

A key advantage of Siamese architectures over other, more traditional, neural networks is their independence of the order of inputs. Feeding both choices as one input into the network can lead to different outputs depending on the order, which is circumvented by having a separate forward-pass through the network for each input. The output for a given input is called the \textbf{embedding} of the input (see Figure \ref{fig:scheme}).  

To compare the different embeddings, Siamese networks often employ the distance between them to model similarities and preferences. The contrastive loss \cite{contrastiveLoss} and triplet loss are common loss functions. 
\begin{equation}
\label{eq:triplet}
    L_{\textrm{triplet}}(\vect{a},\vect{p},\vect{n}) = \max\left(d(\vect{a},\vect{p})-d(\vect{a},\vect{n}) + m,0\right)
\end{equation}

The \emph{triplet loss} (Equation \ref{eq:triplet}) uses an anchor ($\vect{a}$), a positive ($\vect{p}$) and a negative ($\vect{n}$) example. The anchor models the item to compare to, while the positive example $\vect{p}$ is in some manner preferential to the negative example $\vect{n}$. As the loss decreases with decreasing distance between $\vect{a}$ and $\vect{p}$, and with increasing distance between $\vect{a}$ and $\vect{n}$, this means that preferential items are embedded at closer positions in the embedding space than less preferential ones. While this choice is arbitrary, the Euclidian distance $d(\vect{x},\vect{y}) = ||\vect{x} - \vect{y}||_2$. is chosen as the distance metric for this work.
The \emph{margin} $m$ is a parameter of the loss function and controls how far embeddings are pushed away from each other. We used a margin of 1. In preliminary experiments, the exact value of this parameter was not critical for the performance of the method. 

Siamese networks are often used to model the similarity of items. For example, Siamese architectures can compare pictures of individuals and be trained to recognize whether two different images show the same person. In that case, the preference indicates which picture of $p$ and $n$ is more likely to show the same individual as the anchor, therefore modeling similarity between items.

\section{The contextual preference ranking framework for set addition problems}
\label{sec:CPR}

We use Siamese networks differently: instead of item similarity, we model preferences in a contextual, set-based setting, where $\vect{p}$ and $\vect{n}$ are possible additions to an existing anchor set $\vect{a}$.
Formally, this \emph{set addition problem} can be represented as follows:
Given a set of items $\mycards$ modeling the context, and a set of items $\pack$ that represent the current possible choices, select the item $\card^*$ in  $\pack$, which fits the set $\mycards$ best. Formally, if $u(.)$ is an (unknown) utility function that returns an evaluation of a given set of items, then
\begin{equation}
    \card^{*} = \arg \max_{\card \in \pack} u(\mycards \cup \{\card\})
\end{equation}

The learning problem now is to learn the function $u(.)$ from a set of example decisions. We propose to solve this problem by learning contextual preferences of the form 
\begin{equation}
        \left(\card_j \pref \card_k \mid \mycards\right)
\end{equation}
which means that item $\card_j$ is a better addition to set $\mycards$ than $\card_k$. In our application to drafting, all preferences are defined over one-element extensions $\{\card_i\}$. However, in principle, this framework can also be applied if the set $\mycards$ can be extended by arbitrary larger sets of items $\mycards_j$ and $\mycards_k$. For decisions without a context, such as the first pick in a draft, $\mycards = \emptyset$. The distance to the empty set $\emptyset$ can be used as a measure of the general utility of a card.

For training a network with such contextual preference decisions, we employ Siamese networks trained with the triplet loss. While such networks have been previously used for comparing the similarity of items (''anchor object $\vect{a}$ is more similar to object $\vect{p}$ than to object $\vect{n}$''), we use them here in a slightly different setting. The anchor object $\vect{a}$ is a set which needs to be extended with one of two candidate extensions $\vect{p}$ or $\vect{n}$. The training information indicates that $\vect{p}$ is a better extension than $\vect{n}$. This is very different from asking whether $\vect{a}$ is more similar to $\vect{p}$ or $\vect{n}$. For example in card drafting, we seek complementary cards that add to a deck, rather than endlessly duplicating the effect of similar cards picked earlier.

At testing time, pairwise comparisons are not needed, as we can directly evaluate each option in the context of their common anchor.
This is possible because the resulting preferences are transitive w.r.t.\ the given anchor set, i.e., 
\begin{equation*}
(\card_1 \pref \card_2 \mid \mycards)
\land
(\card_2 \pref \card_3 \mid \mycards)
\Rightarrow
(\card_1 \pref \card_3 \mid \mycards)
\end{equation*}
The reason for this is that all objects are embedded with the same embedding network $N$, which always outputs the same signal for the same input, regardless of the position of the item in the comparison.

This Contextual Preference Learning framework is the main contribution of this work, as it introduces a new way of thinking about the Siamese structure. Instead of comparing similar items, we train a preference of items based on a context. To our knowledge, Siamese networks have not previously been used in such a way. In addition, this contextual preference of comparing $\vect{p}$ and $\vect{n}$ with context $\vect{a}$ also differs from comparing $\vect{a}+\vect{p}$ and $\vect{a}+\vect{n}$ as in \emph{RankNet}\cite{RankNet}. We want to emphasize the generality of this framework - it is applicable to model any kind of preference learning problem with a context.

\section{Experimental setup}
\label{sec:experiments}

In this section, we evaluate the framework defined above in the domain of drafting cards in MTG. \footnote{The code used for all experiments can be found at \url{https://github.com/Tibert97/Predicting-Human-Card-Selection-in-Magic-The-Gathering-with-Contextual-Preference-Ranking}}
We define the context $\mycards$ as the set of cards previously chosen by a player and train the networks with pairs of possible card choices $p$ and $n$, where $p$ was chosen by the player and $n$ is another card that was available but not chosen. Therefore, we model that in the human expert's opinion, $p$ fits better into the current set than $n$. When using the network to make a pick decision in a game where we already hold cards $\mycards$, we compute the embedding $N(\mycards)$ and the embeddings $N(\card_i)$ for all possible card choices $\card_i$, then choose the card $\card^*$ with minimal distance to $\mycards$. Due to the nature of how we structure the training in Contextual Preference Ranking, it is important to emphasize that this distance does not model that this card is most similar to $\mycards$. Rather, the distance models how well cards fit into the context, with smaller distances equaling a better fit.

\subsection{Data preparation and exploration}

The \textsl{DraftSim} dataset used in this research consists of 107,949 human drafts from the associated website \cite{b2}. Each draft consists of 24 packs of 15 cards distributed as explained in Section \ref{sec:Drafting}. The dataset includes 2,590,776 separate packs and it contains a total of 265 different cards. It is important to note that those decisions are obtained from a simulator specifically created for drafting. Therefore, the dataset does not contain the playing phase of the game. In addition, the dataset is not tied to a larger \emph{Magic:The Gathering} environment, which means that cards are not associated with a market where cards can be bought or sold. This is important, as otherwise, the physical or digital price of cards may influence decisions.

We train the network on pairs of possible cards in the context of the set of cards that are already held by the player. For each decision to choose the best card from a pack of $k$ cards, $k-1$ training examples are generated for pairing the selected card with each of the $k-1$ other cards in the pack. The \textsl{DraftSim} dataset contains 217,624,680 such training examples. These examples are split 80/20 into training and test data, using the same split as in \cite{b1} to allow a direct comparison.

In order to better understand the 
characteristics of the dataset, we defined two metrics:

\begin{enumerate}[(i)]
   \item The \emph{pick rate} of each individual card $\card$ captures how often the card was selected when being offered. $$p_{\mbox{pick}}(\card) = \frac{\textrm{number of times\ } \card \textrm{\ chosen}}{\textrm{number of times\ } \card \textrm{\ offered}}$$
    \item The \emph{first-pick rate} captures how often a card $\card$ was selected on the very first pick. $$p_{\mbox{firstPick}}(\card) = \frac{\mbox{number of times\ } \card \mbox{\ chosen first}}{\mbox{number of times\ } \card \mbox{\ offered first}}$$
\end{enumerate}

The former metric defines how likely a card is to be chosen over the whole range of the draft, while the second only considers the very first pick. Whether a card is selected first mainly depends on its individual card strength. In contrast, later card choices
are heavily influenced by previously selected cards. In practice, players strongly prefer cards that match their collected colors, as those are most likely to be included in the final deck. 

\begin{figure}
    \includegraphics[width = 0.9 \linewidth, height = 6cm]{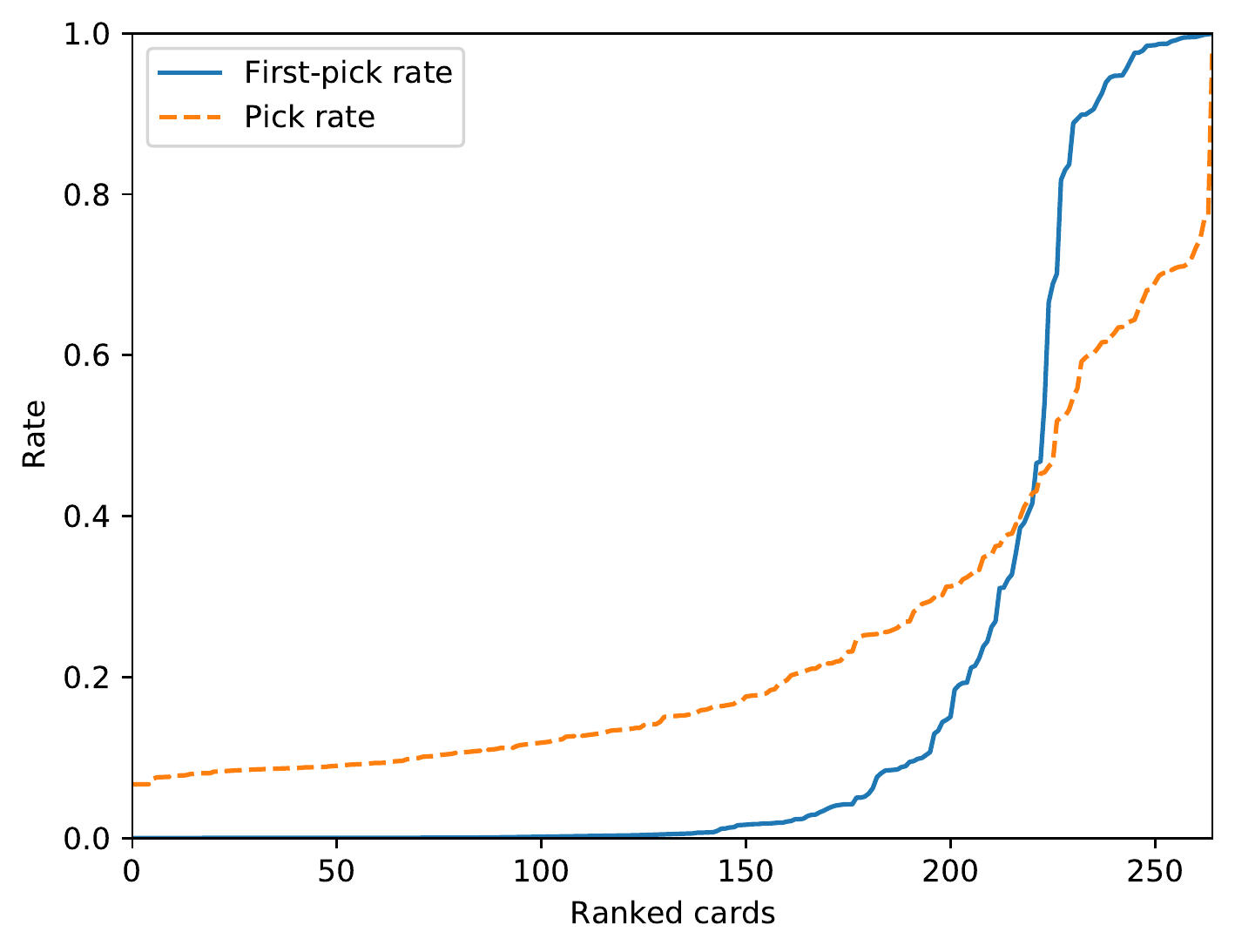}
    \caption{Pick rate of each individual card (higher pick rate equals better card)}
    \label{fig:pickrates}
\end{figure}

Figure \ref{fig:pickrates} demonstrates that recognizing the first picked card is a much easier task than choosing cards later, since the human player's consensus is higher at that point. For the first pick decision, it is possible to simply consult a ranking of available cards \cite{b4,b5,b6}. However, even for this seemingly simple task, rankings are rarely completely unanimous, which underlines the complexity of the domain.

Over the whole draft, all cards will be chosen at some point. For the first pick, the number of reasonable choices is relatively small. Therefore, the first-pick rate drops drastically as cards get weaker, as can be seen from the quick drop of the blue solid line in Figure~\ref{fig:pickrates}. The lowest observed pick rate in the \textsl{DraftSim} set is $0.07$, which is close to the theoretical minimum of $\nicefrac{1}{15} \approx 0.0667$ when a card is always chosen last. However, the lowest first-pick rate in the data set is $0.00001$, which can safely be regarded as a misclick or otherwise unexplainable decision. In contrast, the two highest first-pick rates are 0.9995 and 0.9987, showing clearly that in a vacuum, some cards are clearly regarded as the strongest. The pick-rate differs, as in those decisions the context of already chosen cards is important. There, the two highest pick rates are 0.98 and 0.77. This steep decline occurs, as the card with the highest rate is a colorless one and therefore playable in any deck. The second best however is a white card, which explains why a portion of decisions did not choose that card, as the player was likely already firmly drafting a deck of different colors than white. 

\begin{figure}
    \includegraphics[width = 0.85\linewidth]{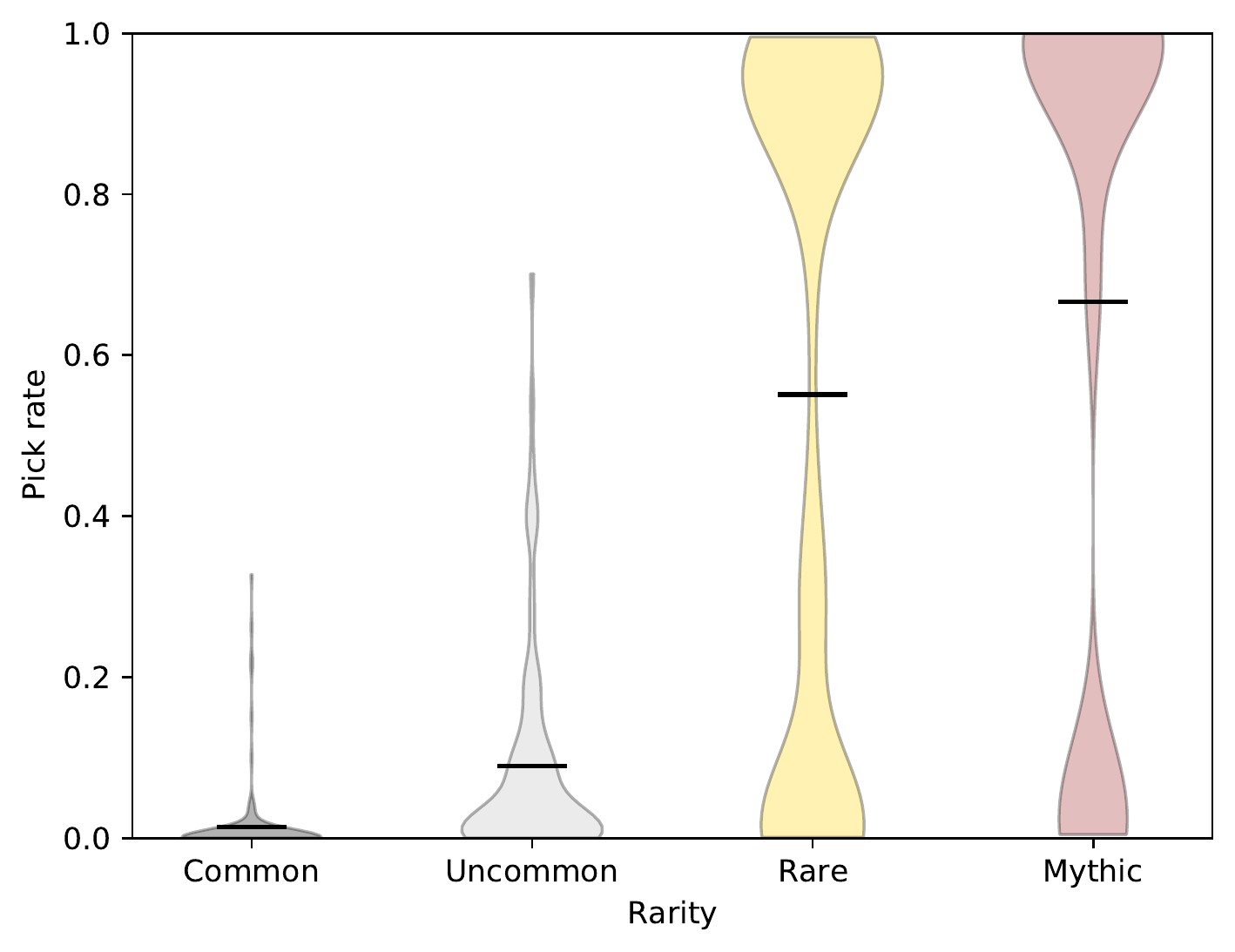}
    \caption{First-pick rate per rarity. Larger area of plot equals more cards of that pack rate. The cards with the highest first-pick rates are all \emph{Rare} and \emph{Mythic}.}
    \label{fig:firstpickrate_rarity}
\end{figure}

Due to the properties of the game, a drafting system for this dataset does not need to be able to compare every card with all others, as a single pack of cards never includes multiple \emph{Rare} or \emph{Mythic} cards (Section \ref{sec:Drafting}). Figure \ref{fig:firstpickrate_rarity} visualizes the density of first-pick rates of cards separated by rarity. There, it is visible that all the very strongest cards in the set are in those two groups. For example, even one of the lesser picked \emph{Rare} cards in the top cluster, with a first-pick rate of about 0.8, is picked more often than any \emph{Common} or \emph{Uncommon} card. However, just choosing \emph{Rares} and \emph{Mythics} whenever possible does not result in an appropriate heuristic. As also seen in Figure \ref{fig:firstpickrate_rarity}, a large number of these cards are also among the least-picked and therefore weakest cards. This is a result of \emph{MTG} having multiple formats. Such cards can be strong within the context of very specific pre-constructed decks but are close to useless in the drafting format.

\subsection{Network Architecture}

\begin{figure}[t]
    \centering\includegraphics[height = 4cm, width = 0.6 \linewidth]{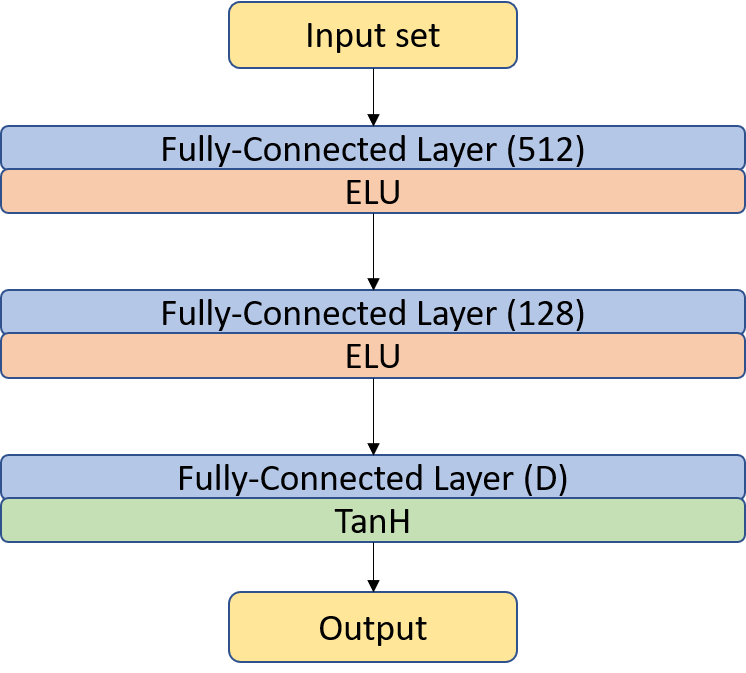}
    \caption{Siamese network \textit{N} architecture}
    \label{fig:siamese}
\end{figure}

This section outlines details of the architecture and training method used for the Siamese network in our experiments. 
The three different inputs $\vect{a}$ (corresponding to the anchor card set $\mycards$), and $\vect{p}$ and $\vect{n}$ (corresponding to the picked card and one of the other cards) are sequentially processed by the network, as shown in Figure~\ref{fig:scheme}. Each forward-pass through the network encodes a set of input cards through multiple fully-connected network layers (Figure \ref{fig:siamese}).
Therefore, each training update consists of three sequential forward passes through the network, followed by the computation of the loss and a backward pass for updating the network parameters.

The embedding network takes a set of cards as input. The input space is 265-dimensional with one dimension representing each possible card. For $\vect{p}$ and $\vect{n}$, the input is a one-hot encoding, while the anchor $\vect{a}$ uses an encoding in which each dimension encodes the number of already chosen cards of that type.
The output of the network is a $D$-dimensional vector of real numbers in the range $[-1,1]$, where $D \geq 1$ is a parameter, which we experimentally evaluate in Section~\ref{sec:results-embedding}. The output vector is the learned embedding of the input set.

Fully-connected layers are linked by exponential linear unit functions (ELU) \cite{b3}. In preliminary experiments, this led to quicker training than rectified linear (RELU) and leaky-RELU activations. We use a learning rate of $0.0001$ and the Adam optimizer with a batch size of $128$. For the output layer, the $\tanh{}$ function was chosen. We do not use batch normalization as it did not provide benefits in our experiments but we use a dropout of $0.5$. Most of those parameters, such as the learning rate, the size of the network, and the optimizer, were not optimized, as reaching the absolute highest performance was not the priority of this work. Rather, we used intuitive parameters, which were comparable to the ones used in previous research \cite{b1}. Performance can likely be enhanced further with a guided search for the optimal parameters.

\section{Results}
\label{sec:results}
In this section, we discuss the performance of our networks for the card selection task and visualize the obtained card embeddings.

\subsection{Card Selection Accuracy}

Our primary goal was to compare our Contextual Preference Ranking framework to the performance of the previous algorithms for this dataset. The best performing algorithm reported in \cite{b1} uses a traditional deep neural network to learn a ranking over all possible cards for a given context. It was trained by directly mapping an encoding of the current set of cards $\mycards$ to a one-hot encoded vector that represents the selected card. Thus, it generated exactly one training example per card pick.
Our Magic draft agent \textsc{SiameseBot} instead learns on pairwise comparisons between the picked card and any other card in the candidate pack $\pack$ and therefore generates $2$ to $14$ training examples from a single pick, depending on the size of $\pack$.

This additional constant factor in the training complexity is to some extent compensated by the fact that we were able to train our network with a much smaller number of training epochs. Due to the large size of the dataset, we split it into 220 sub-datasets. For the results in Figure~\ref{fig:dimension_performance}, only a single epoch of 50 of those datasets was used. We, therefore, used less than $\nicefrac{1}{4}$ of an epoch of the whole dataset, in contrast to 20 epochs of training on the complete dataset in \cite{b1}.

\begin{table}[t]
\caption{Performance of proposed agent to previously seen:\\ $^1$ Heuristic agents \cite{b1} $^2$ Trained agents \cite{b1} $^3$ This work}
\centering
\resizebox{0.7\linewidth}{!}{%
\begin{tabular}{|l|c|l|}
\hline
\textbf{Agent} & \textbf{MTTA (\%)} & \textbf{MTPD} \\ \hline\hline
RandomBot$^1$    & 22.15 & NA     \\ 
RaredraftBot$^1$ & 30.53 & 2.62   \\ 
DraftsimBot$^1$  & 44.54 & 1.62   \\ \hline
BayesBot$^2$     & 43.35 & 1.74   \\ 
NNetBot$^2$      & 48.67 & 1.48   \\ \hline 
SiameseBot$^3$, D=2   & \textbf{53.69} & \textbf{0.98} \\ 
SiameseBot$^3$, D=256   & \textbf{83.78} & \textbf{0.2476} \\ \hline
\end{tabular}%
}
\label{tab:performance_comp}
\end{table}

Following \cite{b1} we report two measures: the \textit{mean testing top-one accuracy (MTTA)} is the percentage of cases in which the network chooses the correct card in the pack. The \emph{mean testing pick distance (MTPD)} is how far away the correct pick is from the chosen card when ranking all possible choices. In both of those metrics, with embedding dimension $D = 256$, we achieve substantially improved results on the dataset, as can be seen from Table~\ref{tab:performance_comp}.

This strong increase in performance suggests that our Contextual Preference Ranking approach works well for this domain. Furthermore, the proposed approach is completely domain-agnostic. Apart from having a fixed one-hot encoding of each card, we do not provide the network any other information about the game or the cards. This leads us to speculate that our method will likely work well for other contextual decision-making problems.

\begin{figure}
    \centering
    \includegraphics[width = 0.85 \linewidth]{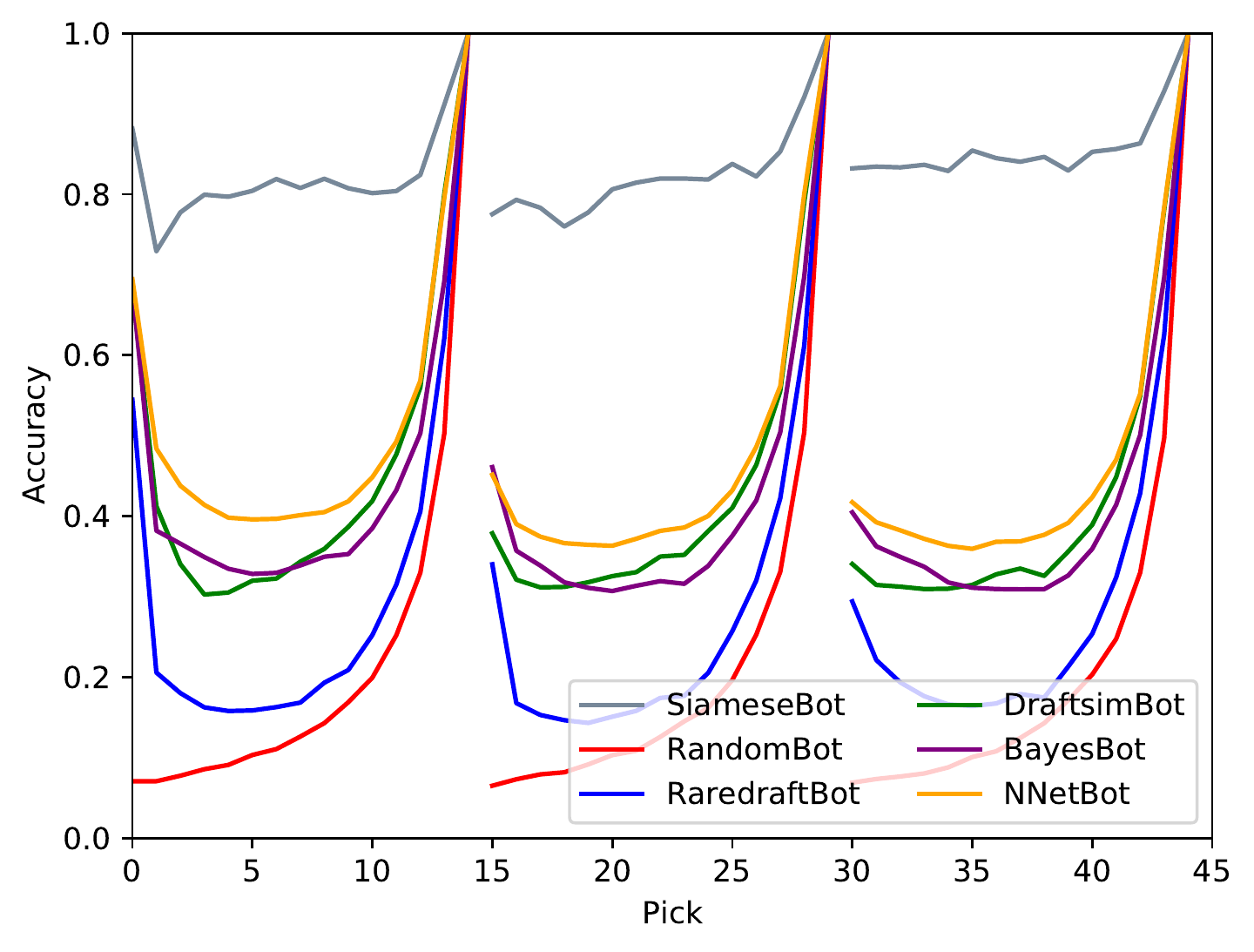}
    \caption{Accuracy of pick-prediction per number of already chosen cards. \textsc{SiameseBot} performs much better than all other methods and remains a more stable accuracy in the middle of the packs.}
    \label{fig:accuracy_per_pick}
\end{figure}

\subsection{Draft Analysis}
We also compare the performance of our proposed network over the course of the whole draft. Since already chosen cards strongly influence the current decision, we explore whether a growing set of chosen cards influences the accuracy of picks.  Figure~\ref{fig:accuracy_per_pick} shows the accuracy of our \textsc{SiameseBot} and those reported in \cite{b1} over the three consecutive picking rounds with 15 cards each. We clearly see that our method generally provides substantially more accurate decisions. Interestingly, the accuracy of picks does not show the same performance curve as previous methods. Those methods have U-shaped curves and are more accurate at the start and end of each pack. Our method remains a relatively stable quality throughout most of the pack. The worst accuracy for our \textsc{SiameseBot} occurs at pick number 2, which is an interesting observation. A possible reason for this may be, that the embedding of the context with a single card, and the embedding of a possible choice card itself are the same, which could lead to problems there.

\subsection{Visualization}
Finally, we can use the resulting embedding of cards to visualize the decision process of the network. Since embeddings for cards are constant, the card selection decision is solely determined by the embedded representation of the anchor. 

Visualizing the embedding of single cards aids in the understanding of the decision process of the network. As the embedding is 256-dimensional, we use t-SNE \cite{b7}, a stochastic algorithm to reduce the dimensionality of data points, to plot the embedding in two dimensions. We graph all cards in their respective color. Cards of exactly two colors are shown with one color as their border and the other as the filling. Purple is used for colorless cards, and gold for cards of more than two colors. The empty set is shown as the anchor in the middle, which corresponds to visualizing the first pick.

\begin{figure}
    \centering
    \includegraphics[width = 0.85 \linewidth]{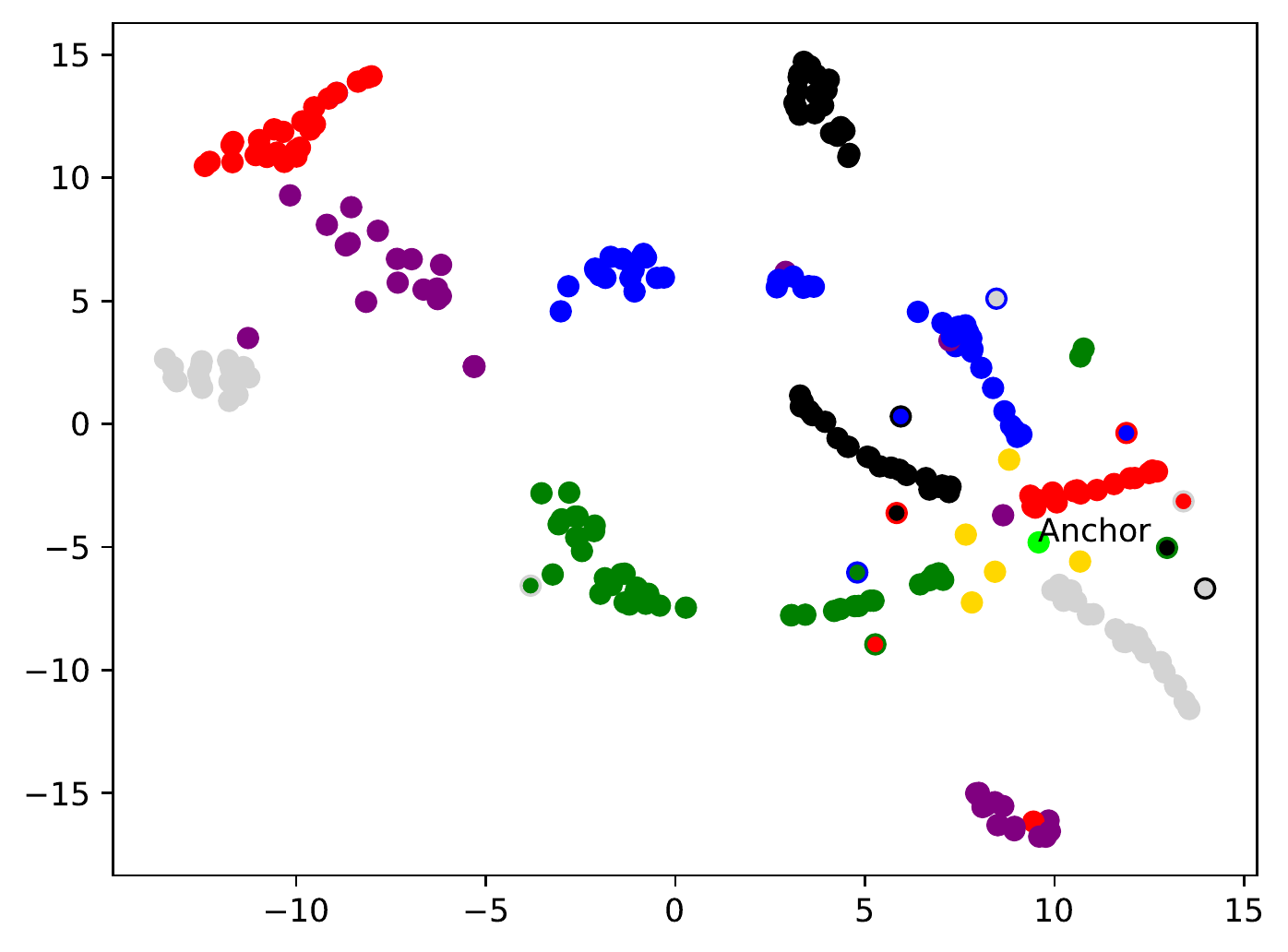}
    \caption{2D visualization of embedded cards with the anchor as the empty card set and colors matching the card colors in the game. Clear clusters of equal-colored cards are visible, although the network did not receive this information.}
    \label{fig:coloured_embedding}
\end{figure}

Although the network did not receive any information about the color of cards, Figure~\ref{fig:coloured_embedding} shows clear clusters with cards of the same color. Within each cluster, cards seem to be roughly linearly ordered, where cards closer to the empty set are stronger in a vacuum.  This leads to a star-like structure. Between clusters, single multicolored cards are visible, which correspond to multicolored cards of the two adjacent colors.

However, some clusters outside the star structure are visible. As the dimensionality of the embedding was reduced drastically, giving an accurate explanation of those is far from trivial. Firstly, as t-SNE is stochastic, the resulting 2-dimensional representation changes in different iterations. Therefore, some clusters sometimes seem to be connected, while they are disconnected in other iterations. Figure \ref{fig:coloured_embedding} shows two distinct black clusters, but in other iterations with the same parameters, the structure of those is more connected. We also use k-means to cluster the data in its original space. There, the far-seeming white points are sometimes clustered together, which further strengthens the explanation that some structures are artifacts of t-SNE. This is even more drastic when taking different hyperparameters into account.

Finally, we tested how the observable distances to the anchor in Figure~\ref{fig:coloured_embedding} correlate to the real distances in the 265-dimensional embedding space. Those two measures achieve a Kendall's Tau correlation of 0.6243, which means they are strongly correlated, showing that Figure~6 still gives good intuition about the decision process of the network. When investigating the correlation further, the loss in correlation mainly comes from singular points and clusters which have drastically different distances in the two embedding spaces, while clusters themselves seem to achieve a similar distance. 

\subsection{Sensitivity to Embedding Dimension}
\label{sec:results-embedding}

\begin{figure}
    \centering
    \includegraphics[width = 0.8\linewidth]{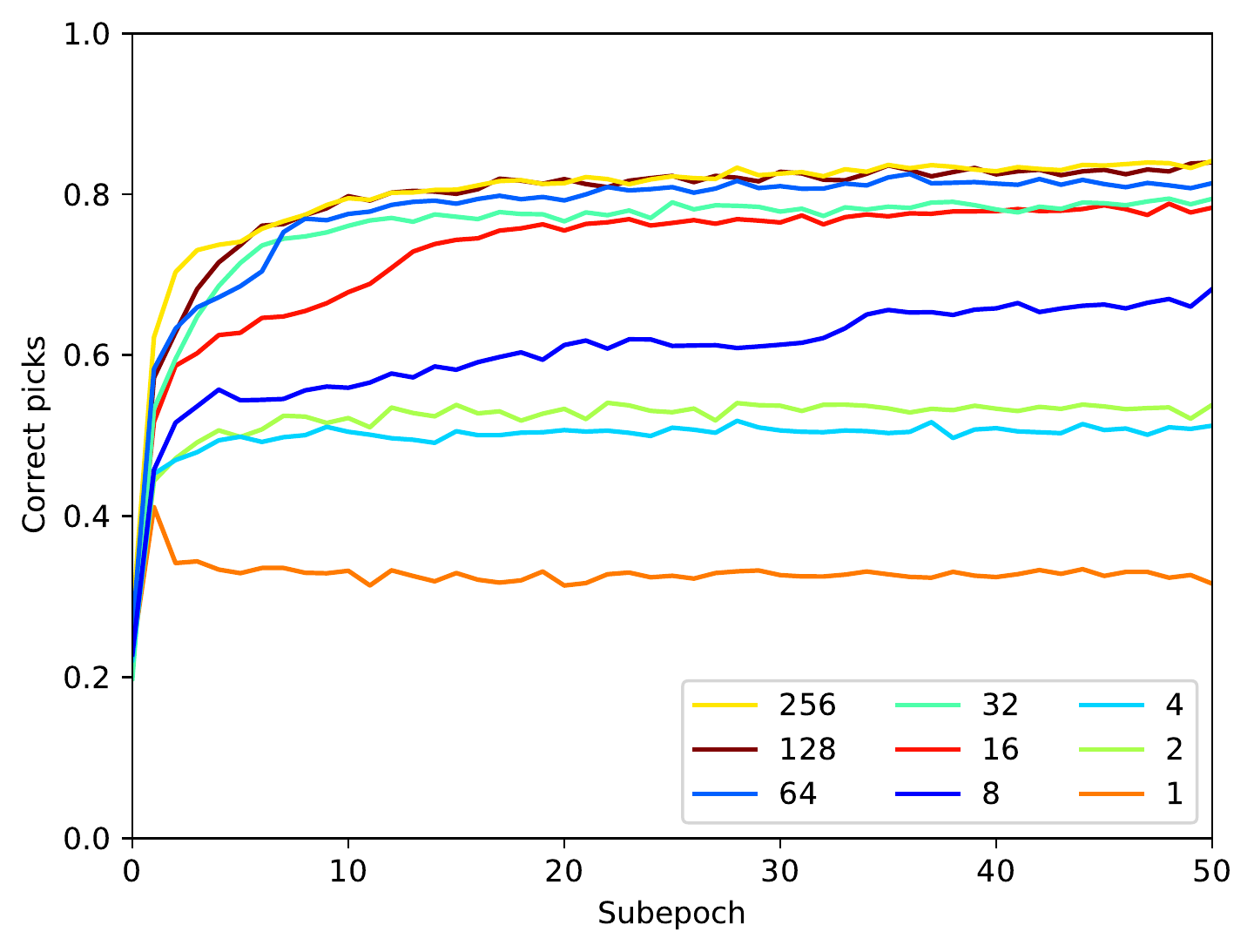}
    \caption{Influence of $D$ on performance. Increasing $D$ leads to higher accuracies until $D \approx 128$.}
    \label{fig:dimension_performance}
\end{figure}

The embedding dimension $D$ is the most important hyperparameter of the proposed method. Figure \ref{fig:dimension_performance} shows the learning curves for different choices of $D$. Increasing $D$ leads to strong improvements in network accuracy up to about $D=32$. After this, diminishing returns set in and the performance stops improving at $D=128$. However, even with $D=2$, the Siamese network achieves an accuracy of 53.69\%, which is higher than the 48.67\% of the best previous method \textsc{NNetBot} (see Table \ref{tab:performance_comp}). 

\section{Discussion}

With the proposed approach, we achieve much higher accuracies than previously reported, while using a much smaller amount of training. Besides, the embedding of cards and decks provides valuable information about the dataset without any added computational effort. The fact that cards of the same color are clustered together is intuitive and further confirms the validity of our approach. Embedding a different dataset would likely look vastly different, for example, a set where specific colors are more likely to be drafted together. One surprising finding was that the embedding was not intuitively perfect. A few outside clusters of cards  (Figure \ref{fig:coloured_embedding}) seem to be rated drastically worse than the main structure around the anchor. A possible explanation is that those clusters contain weak cards which are chosen near the end. This would still cluster them together, as they are only chosen in accordance with the color of the deck while being exceptionally far away from the empty set since they are never chosen as the first pick. It however could also just be an artifact of the dimensionality reduction of the data. Another reassuring finding is that two-colored cards lie between their two colors, as those cards are equally relevant for both.

We can use the resulting embedding to construct a rating of all cards by computing their distances to the empty set. Interestingly, this differs from expert rankings. We compare this resulting ranking to two expert opinions in Table \ref{tab:ranking_comp}. The last column ranks the cards based on how often they were first picked in the dataset (compare Figure \ref{fig:pickrates}). The rarity of each card is encoded behind its name as either \emph{Uncommon (U)}, \emph{Rare (R)} or \emph{Mythic (M)}.

From this, a few stark differences are immediately obvious. While \emph{Ajani, Adversary of Tyrants}, \emph{Djinn of Wishes} and \emph{Leonin Warleader} are rated similarly to the experts, the extraordinarily high rating of \emph{Goblin Trashmaster} is surprising. Note that the rates for the top cards are very similar, e.g. 99.94\% for \emph{Ajani, Adversary of Tyrants} and 99.87\% for \emph{Resplendent Angel}, and those \emph{Rare} and \emph{Mythic} cards are never in direct competition due to the composition of the decks. This can make a correct ranking very hard for the network. We can also observe from the Siamese ranking that, surprisingly, four of the top five cards are \emph{Rare}. We speculate that this is due to the fact that \emph{Rare} cards occur more frequently in the dataset and the training sees more positive examples of these. It is possible to combat this by oversampling mythic examples, or by adding features to the cards, but this would stand in contrast to the domain-agnostic approach chosen. 
While the high ranking of \emph{Goblin Trashmaster} is unusual, the network has, however, made a precise estimation about a hard-to-rate card, \emph{Murder}. This is by far the best \emph{Uncommon} card in the set. It is ordered at rank 9 and 21 by CFB and DraftSim respectively. Our Siamese network ranks it at 12, although its first-pick rate is only 39.

\begin{figure}[t]
    \centering
    \includegraphics[width = 0.82 \linewidth]{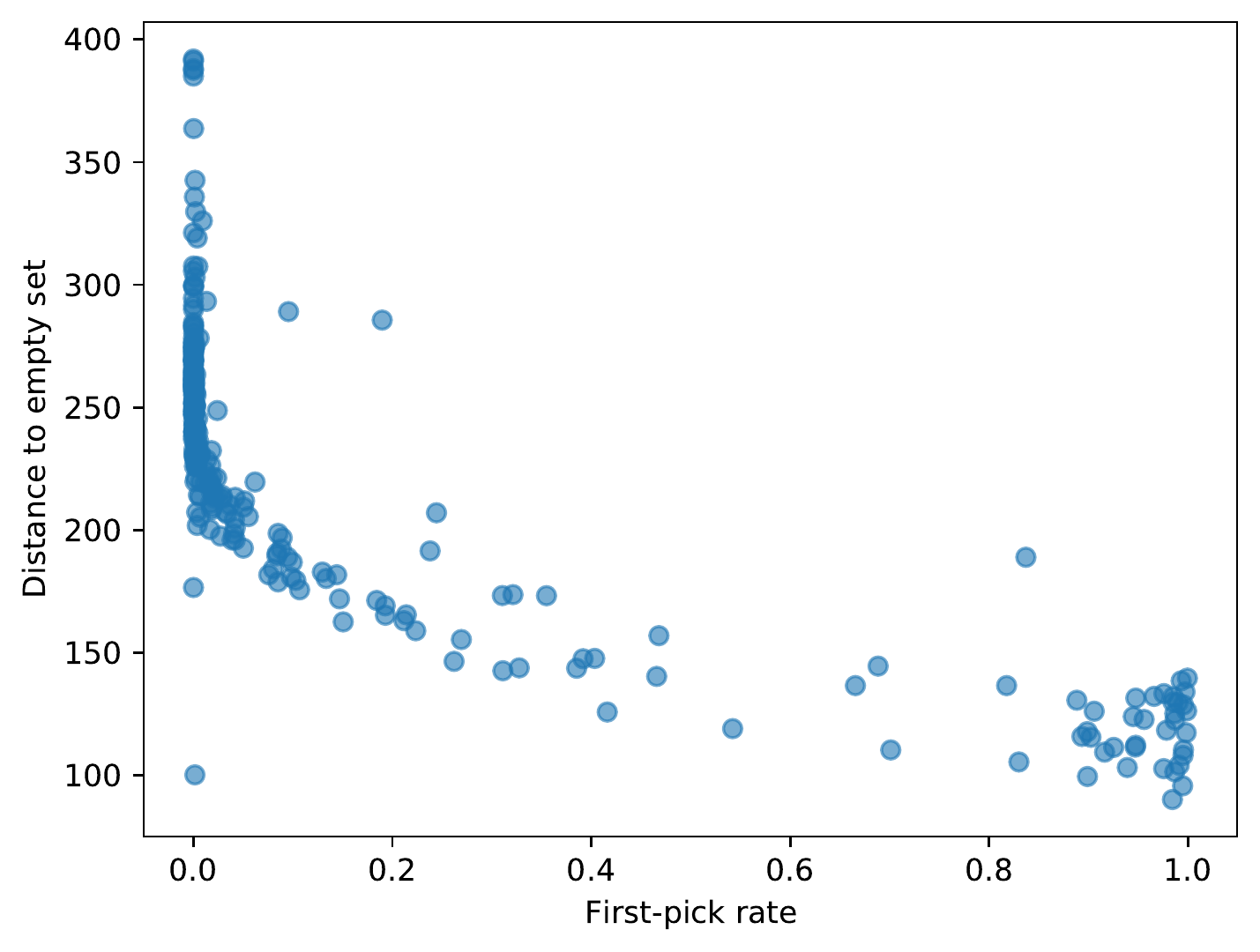}
    \caption{Correlation between first-pick rate and distance. Cards with a higher first-pick rate are embedded closer to the empty set. Kendall's Tau = 0.74}
    \label{fig:fpr_distance}
\end{figure}

\begin{table}[t]
\caption{Rankings of Firstpicks of the proposed method compared to expert opinions. FPR = Firstpickrate}
\label{tab:ranking_comp}
\resizebox{\linewidth}{!}{%
\begin{tabular}{|l|c|c|c|c|}
\hline
\textbf{Card} & \textbf{Siamese} & \textbf{Expert 1 \cite{b6}} & \textbf{Expert 2 \cite{b4}} & \textbf{FPR} \\ \hline\hline
Spit Flame (R)                 & 1  & 18 & 22  & 17 \\ 
Leonin Warleader (R)            & 2  & 15 & 4   & 8  \\ 
Goblin Trashmaster (R)         & 3  & 51 & 112 & 32 \\ 
Ajani, Adversary of Tyrants (M) & 4  & 7  & 5   & 1  \\ 
Djinn of Wishes (R)             & 5  & 14 & 6   & 14 \\ 
Tezzeret, Artifice Master (M)   & 20 & 1  & 2   & 3  \\ 
Resplendent Angel (M)           & 30 & 9  & 1   & 2  \\ 
Murder (U)           & 12 & 21  & 9   & 39  \\ \hline
\end{tabular}%
}
\end{table}
To further visualize correlations between the network predictions and the underlying data, we plot the first-pick rate of cards against the distance to the empty set in Figure~\ref{fig:fpr_distance}, showing a strong correlation with a Kendall rank correlation coefficient of 0.74.
The main difference between these two statistics is that the distance is much smoother than the first-pick rate, which decreases rapidly for weaker cards. The first-pick rate is only subject to binary choices, i.e., $\card_1 \pref \card_2$\, without giving any weight to how close the decision between those cards was. Due to the training with more than just the first picks, the embedding distance is a smoother measure of how strong the card is according to the network. 

Finally, we can use the embedding to extract meta-information about this dataset. For example, the Siamese network seems to strongly favor the colors red and white, as it rates four white and five red cards higher than the best green one. 

\section{Conclusion}
We showed that the proposed method of using a Siamese network to model preferences in the context of drafting cards in \emph{Magic: The Gathering} worked well and vastly outperformed previous results. Compared to \cite{b1}, we report an increase in accuracy by more than 56\%, while also decreasing the pick distance by more than 83\%. 
Even when our network makes an incorrect choice, the network ranks the correct choice very high. In addition to this performance, we show that the resulting embedding makes intuitive sense. It can be used to learn further from the dataset, apart from only using it for draft predictions. For this dataset, we were able to create absolute rankings of cards and could speculate which colors \textsc{SiameseBot} prefers.

With this first implementation of a contextual preference ranking framework, we showed that Siamese networks work well for adding items to an existing set. We want to reemphasize that while this is the first practical test of this framework, there is no reason to believe that the success is limited to this particular setting. We did not incorporate any domain information into \textsc{SiameseBot} beyond the ID of cards used to encode the input. Therefore, we speculate that our proposed framework will work well for other problems where preference has to be modeled in a context.

\section{Future work}
In order to further test the generality of this approach in other domains, more work with other datasets is required. One possible area for future work is sequential team-building in a MOBA game. It could also be possible to extend this approach beyond sequential decision-making. An example is a game where decks are played against each other, and the context is the intersection of both decks, with positive and negative examples taken from the remaining cards of the winning and losing deck respectively. This may introduce a lot of noise into the training, as winning or losing with a deck is subject to a multitude of factors besides the chosen cards, but may extend the method to a larger variety of domains. 

There is potential to use this method not only for pre-game decision-making but for game playing as well. Given a dataset of expert moves in a game, we can model the anchor as the current game state, and the chosen and one not chosen move as positive and negative examples. A concern with all of those ideas however is the fact that we are solely training on human expert examples, which provides an upper limit on how well this can perform in a general context. To circumvent this, one could also generate datasets on self-play games as part of an agent training loop, as in AlphaZero \cite{Alpha} and similar approaches.
For further improving performance within MTG, we could build refined architectures that use meta-information and history, which allows inferences about opponent strategies and color choices which are used by strong human players. We could also try to train separate networks for specific numbers of already chosen cards, especially for the case of 1 chosen card where performance is worst overall.

\medskip
\noindent\small
\textbf{Acknowledgements}
We thank the authors of \cite{b1} for making the data publicly available and for sharing their experimental data, and Johannes-Kepler Universit\"at Linz for supporting M\"uller's sabbatical stay through their Research Fellowship program.

\end{document}